\documentclass[journal]{IEEEtran}
\ifCLASSINFOpdf
\else
\fi
\usepackage{lipsum}
\usepackage{xcolor}
\usepackage{todonotes}
\usepackage{comment}
\usepackage{arydshln}

\usepackage{times}
\usepackage{epsfig}
\usepackage{graphicx}
\usepackage{amsmath}
\usepackage{amssymb}
\usepackage[breaklinks=true,bookmarks=false]{hyperref}
\usepackage{url}
\usepackage{algorithm}
\usepackage{algpseudocode}
\usepackage{listings}
\usepackage{color}
\usepackage{setspace}
\usepackage{courier}
\usepackage{url}
\usepackage{booktabs} % for the tables
\usepackage{multirow,bigdelim}
\usepackage[switch]{lineno}
\usepackage{caption}
\newcommand\blue[1]{\color{black}#1}
\newcommand{\norm}[1]{\left\lVert#1\right\rVert}

\begin{document}
\title{Anatomically Constrained Neural Networks (ACNN): Application to Cardiac Image Enhancement and Segmentation}

\author{Ozan Oktay, Enzo Ferrante, Konstantinos Kamnitsas, Mattias Heinrich, Wenjia Bai, Jose Caballero, Stuart Cook, Antonio de Marvao, Timothy Dawes, Declan O'Regan, Bernhard Kainz, Ben Glocker, and Daniel Rueckert
\thanks{(I) O. Oktay, E. Ferrante, K. Kamnitsas, W. Bai, J. Caballero, B. Kainz, B. Glocker, and D. Rueckert are with Biomedical Image Analysis Group, Imperial College London, SW7 2AZ London, U.K.}
\thanks{(II) S. A. Cook, A. de Marvao, T. Dawes, and D. P. O'Regan are with MRC Clinical Sciences Centre (CSC), London W12 0NN, U.K.}
%\thanks{(III) M. Heinrich is with the Institute of Medical Informatics, University of L\"ubeck, 23538 L\"ubeck, Germany.}% <-this % stops a space
}


\markboth{PUBLISHED IN IEEE TRANSACTIONS ON MEDICAL IMAGING, AUG 2017}%
{Shell \MakeLowercase{\textit{et al.}}: Bare Demo of IEEEtran.cls for IEEE Journals}



\maketitle

\begin{abstract}
Incorporation of prior knowledge about organ shape and location is key to improve performance of image analysis approaches. In particular, priors can be useful in cases where images are corrupted and contain artefacts due to limitations in image acquisition.
The highly constrained nature of anatomical objects can be well captured with learning based techniques. However, in most recent and promising techniques such as CNN based segmentation it is not obvious how to incorporate such prior knowledge. State-of-the-art methods operate as pixel-wise classifiers where the training objectives do not incorporate the structure and inter-dependencies of the output. To overcome this limitation, we propose a generic training strategy that incorporates anatomical prior knowledge into CNNs through a new regularisation model, which is trained end-to-end. The new framework encourages models to follow the global anatomical properties of the underlying anatomy (\emph{e.g.} shape, label structure) via learnt non-linear representations of the shape. We show that the proposed approach can be easily adapted to different analysis tasks (\emph{e.g.} image enhancement, segmentation) and improve the prediction accuracy of the state-of-the-art models. The applicability of our approach is shown on multi-modal cardiac datasets and public benchmarks. Additionally, we demonstrate how the learnt deep models of 3D shapes can be interpreted and used as biomarkers for classification of cardiac pathologies
\end{abstract}

\begin{IEEEkeywords}
Shape Prior, Convolutional Neural Network, Medical Image Segmentation, Image Super-Resolution
\end{IEEEkeywords}

% \fi
\IEEEpeerreviewmaketitle

\section{Introduction}

Image segmentation techniques aim to partition an image into meaningful parts which are used for further analysis. The segmentation process is typically driven by both the underlying data and a prior on the solution space, where the latter is useful in cases where the images are corrupted or contain artefacts due to limitations in the image acquisition. For example, bias fields, shadowing, signal drop-out, respiratory motion, and low-resolution acquisitions are the few common limitations in ultrasound (US) and magnetic resonance (MR) imaging. 

Incorporating prior knowledge into image segmentation algorithms has proven useful in order to obtain more accurate and plausible results as summarised in the recent survey \cite{nosrati2016incorporating}. Prior information can take many forms: boundaries and edge polarity \cite{chen2017dcan}; shape models \cite{cootes1995combining, davatzikos2003hierarchical}; topology specification; distance prior between regions; atlas models \cite{bai2013probabilistic}, which were commonly used as a regularisation term in energy optimisation based traditional segmentation methods (e.g. region growing). In particular, atlas priors are well suited for medical imaging applications since they enforce both location and shape priors through a set of annotated anatomical atlases. Similarly, auto-context models \cite{tu2010auto} have made use of label and image priors in segmentation, which require a cascade of models. 

\begin{figure}[t]
	\centering
	\includegraphics[width=.5\textwidth]{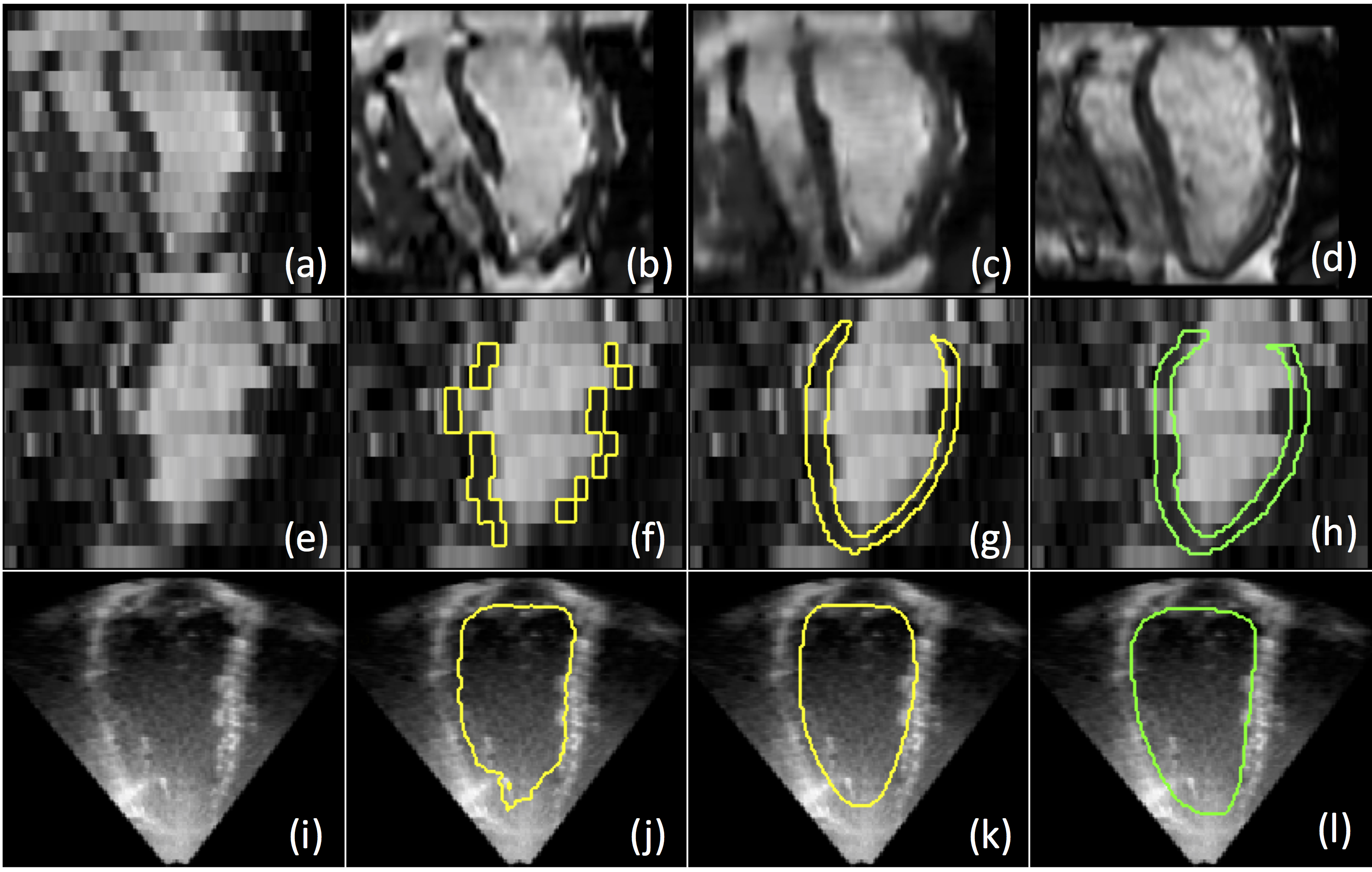}
	\caption{Results for cardiac MR super-resolution (SR) (top), MR segmentation (middle), and ultrasound (US) segmentation (bottom). From left to right, we show the input image, a state-of-the-art competing method,  the proposed result, and the ground-truth. (a) Stack of 2D MR images with respiratory motion artefacts, (b) SR based on CNNs \cite{oktay2016multi}, (c) the proposed ACNN-SR, (d) ground-truth high-resolution (HR) image, (e) low resolution MR image, (f) 2D segmentation resulting in blocky contours \cite{tran2016fully}, (g) 3D sub-pixel segmentation from stack of 2D MR images using ACNN, (h) manual segmentation from HR image, (i) input 3D-US image, (j) FCN based segmentation \cite{chen2016iterative}, (k) ACNN, and (l) manual segmentation.}
	\label{fig:ClinicalMotivation}
\end{figure}

In the context of neural networks (NNs), {\blue early work on shape analysis has focused on learning generative models through deep Boltzmann Machines (DBMs), namely ShapeBM \cite{eslami2012SBM} that uses a form of DBM with sparse pixel connectivity. Follow-up work in \cite{chen2013deep,eslami2012generative} has demonstrated the application of DBMs to binary segmentation problems in natural images containing vehicles and other types of objects. However, fully connected DBM for images require a large number of parameters and consequently model training may become intractable depending on the size of images. For this reason, convolutional deep belief nets \cite{wu20153d} were recently proposed for encoding shape prior information. Besides variational models, cascaded convolutional architectures \cite{li2016iterative,ravishankarlearning} have been shown to discover priors on shape and structure in label space without any a priori specification. However, this comes at the cost of increased model complexity and computational needs.}

{\blue In the context of medical imaging and neural networks, anatomical priors have not been studied in much depth, particularly in the current state-of-the-art segmentation techniques \cite{kamnitsas2017efficient, ronneberger2015u, chen2016iterative, ravishankar2017joint}.} Recent work has shown simple use cases of priors through adjacency \cite{bentaieb2016topology} and boundary \cite{chen2017dcan} conditions. Inclusion of priors in medical imaging could potentially have much more impact compared to their use in natural image analysis since anatomical objects in medical images are naturally more constrained in terms of their shape and location.

As explained in a recent NN survey paper \cite{litjens2017survey}, the majority of the classification and regression models utilise a pixel-level loss function (\emph{e.g.} cross-entropy or mean square error) which does not fully take into account the underlying semantic information and dependencies in the output space (e.g. class labels). In this paper, we present a novel and generic way to incorporate global shape/label information into NNs. The proposed approach, namely anatomically constrained neural networks (ACNN), is mainly motivated by the early work on shape priors and image segmentation, in particular PCA based statistical \cite{cootes1995combining} and active shape models \cite{davatzikos2003hierarchical}. Our framework learns a non-linear compact representation of the underlying anatomy through a stacked convolutional autoencoder \cite{masci2011stacked} and enforces network predictions to follow the learnt statistical shape/label distributions. In other words, it favours predictions that lie on the extracted low dimensional data manifold. {\blue More importantly, our approach is independent of the particular NN architecture or application; it can be combined with any of the state-of-the-art segmentation or super-resolution (SR) NN models and potentially improve its prediction accuracy and robustness without introducing any memory or computational complexity at inference time.} Lastly, ACNN models, trained with the proposed prior term which acts as a regulariser, remove the need for post-processing steps such as conditional random fields \cite{lafferty2001conditional} which are often based on heuristics parameter tuning. In ACNN, the regularisation is part of the end-to-end learning which can be a great advantage.

The proposed global training objective in SR corresponds to a prior on the space of feasible high-resolution (HR) solutions, which is experimentally shown to be useful since SR is an ill-posed problem. Similar modifications of the objective function during training have been introduced to enhance the quality of natural images, such as perceptual \cite{johnson2016perceptual} and adversarial \cite{ledig2016photo} loss terms, which were used to synthesise more realistic images in terms of texture and object boundaries. In the context of medical imaging, our priors enforce the synthesised HR images to be anatomically meaningful while minimising a traditional image reconstruction loss function. 
\vspace{-2 mm}
\subsection{Clinical Motivation}

Cardiac imaging has an important role in diagnosis, pre-operative planning, and post-operative management of patients with heart disease. Imaging modalities such as US and cardiac MR (CMR) are widely used to provide detailed assessment of cardiac function and morphology. Each modality is suitable for particular clinical use cases; for instance, 2D-US is still the first line of choice due to its low cost and wide availability, whereas, CMR is a more comprehensive modality with excellent contrast for both anatomical and functional evaluation of the heart \cite{karamitsos2009role}. Similarly, 3D-US is recommended over the use of 2D-US since it has been demonstrated to provide more accurate and reproducible volumetric measurements \cite{lang2012eae}. 

Some of the standard clinical acquisition protocols in 3D-US and CMR still have limitations in visualising the underlying anatomy due to imaging artefacts (\emph{e.g.} cardiac motion, low slice resolution, lack of slice coverage \cite{petersen2016uk}) or operator-dependent errors (\emph{e.g.} shadows, signal drop-outs). In the clinical routine, these challenges are usually tackled through multiple acquisitions of the same anatomy and repeated patient breath-holds leading to long examination times. Similar problems have been reported in large cohort studies such as the UK Biobank \cite{petersen2016uk}, which leads to inaccurate quantitative measurements or even the discarding of acquired images. As can be seen in Fig. \ref{fig:ClinicalMotivation}, the existing state-of-the-art convolutional neural network (CNN) approaches for segmentation \cite{tran2016fully, chen2016iterative} and image enhancement \cite{oktay2016multi} tasks perform poorly when the input data is not self-consistent for the analysis. For this reason, incorporation of prior knowledge into cardiac image analysis could provide more accurate and reliable assessment of the anatomy, which is shown in the third column of the same figure. Most importantly, the proposed ACNN model allows us to perform HR analysis via sub-pixel feature maps generated from low resolution (LR) input data even in the presence of motion artefacts. Using the proposed approach we can perform full 3D segmentation without explicit motion correction and do not have to rely on LR slice-by-slice 2D segmentation.

We demonstrate the applicability of the proposed approach for cine stacks of 2D MR and 3D-US datasets composed of $1200$ and $45$ cardiac image sequences respectively. We show that the proposed segmentation and SR models become more robust against imaging artefacts mentioned earlier which is underlined by our state-of-the-art results on the MICCAI'14 CETUS public benchmark \cite{bernard2016standardized}.  We also demonstrate that the lower dimensional representations learnt by the proposed ACNN can be useful for classification of pathologies such as dilated and hypertrophic cardiomyopathy, and it does not require point-wise correspondence search between subjects as in \cite{shakeri2016deep}. For the evaluation, the MICCAI'17 AC/DC classification benchmark was used. In that regard, the proposed method is not only useful for image enhancement and segmentation but also for the study of anatomical shape variations in population studies and their associations with cardiac related pathologies. 

\begin{figure*}[t]
	\centering
	\includegraphics[width=.75\textwidth]{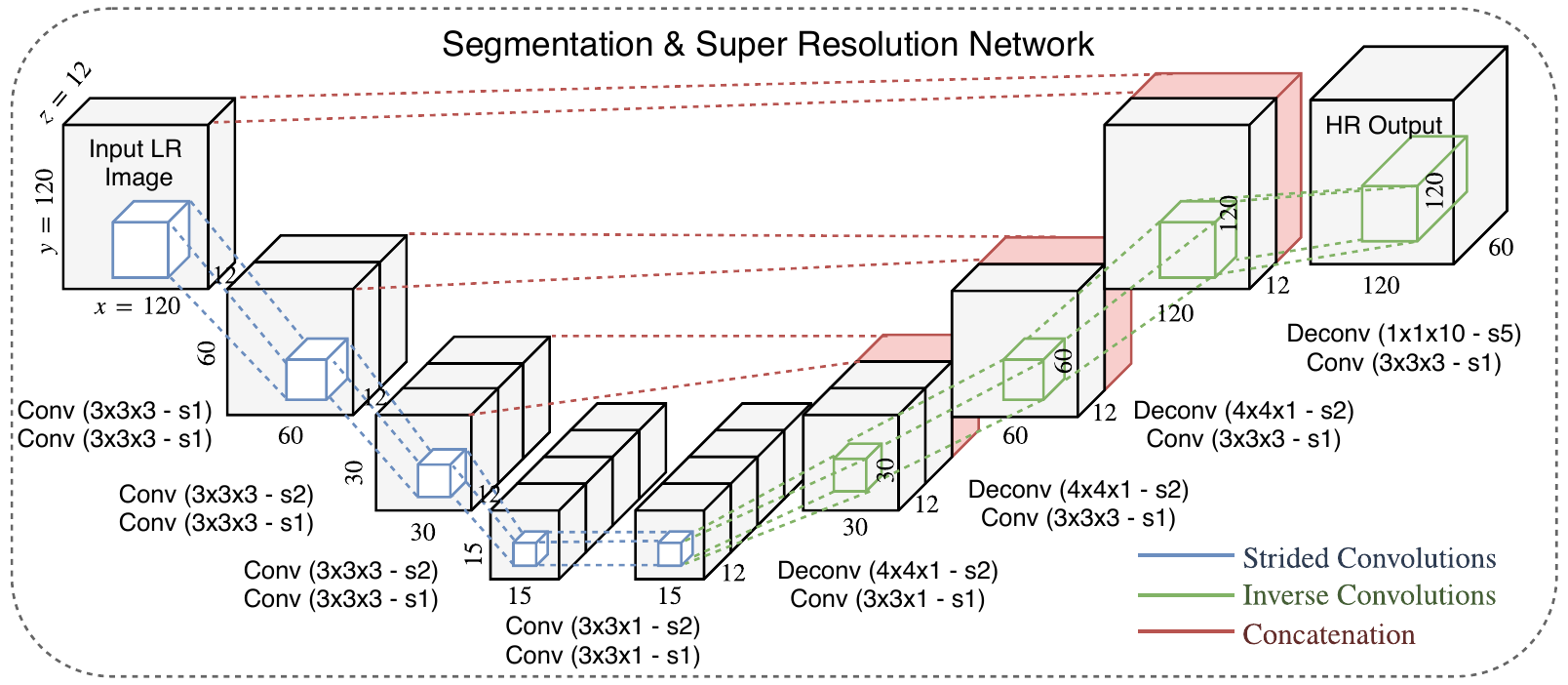}
	\caption{Block diagram of the baseline segmentation (Seg) and super-resolution (SR) models which are combined with the proposed T-L regularisation block (shown in Fig. \ref{fig:ae_tl_network}) to build the ACNN-Seg/SR frameworks. In SR, the illustrated model extracts SR features in low-resolution (LR) space, which increases computational efficiency. In segmentation, the model achieves sub-pixel accuracy for given LR input image. The skip connections between the layers are shown in red.}
	\label{fig:seg_sr_networks}
\end{figure*}

\subsection{Contributions}
In this study, we propose a generic and novel technique to incorporate priors on shape and label structure into NNs for medical image analysis tasks. In this way, we can constrain the NN training process and guide the NN to make anatomically more meaningful predictions, in particular in cases where the input image data is not informative or consistent enough (\emph{e.g.} missing object boundaries). More importantly, to the best of our knowledge, this is one of the earliest studies demonstrating the use of convolutional autoencoder networks to learn anatomical shape variations from medical images. 

The proposed ACNN model is evaluated on multi-modal cardiac datasets from MR and US. Our evaluation shows: (I) A sub-pixel cardiac MR image segmentation approach that, in contrast to previous CNN approaches \cite{tran2016fully,avendi2016combined}, is robust against slice misalignment and coverage problems; (II)  An implicit statistical parametrisation of the left ventricular shape via NNs for pathology classification; (III) An image SR technique that extends previous work \cite{oktay2016multi} and that is robust against slice misalignments; our approach is computationally more efficient than the state-of-the-art SR-CNN model \cite{oktay2016multi} as the feature extraction is performed in the low-dimensional image space. (IV) Last, we demonstrate state-of-the-art 3D-US cardiac segmentation results on the CETUS'14 Benchmark.

\begin{comment}
-\\
Image segmentation papers:
Multi-atlas Segmentation \cite{bai2013probabilistic}
Survey Paper on Deep Learning \cite{litjens2017survey}
DeepMedic \cite{kamnitsas2017efficient}, U-Net \cite{ronneberger2015u}, FCN \cite{long2015fully}
GANs in segmentation \cite{luc2016semantic}
SegNet \cite{badrinarayanan2015segnet}
Ultrasound Seg Paper \cite{chen2016iterative}
\\

Autoencoder papers:
TL-network paper by Girdhar \cite{girdhar2016learning}\\
Convolutional Autoencoder Network by Schmidhuber\\
Shape Learning with Autoencoder \cite{sharma2016vconv}
\\

Super-resolution papers:
Perceptual Losses for SR \cite{johnson2016perceptual}
Application of GANs on SR \cite{ledig2016photo}
CMR SR paper \cite{oktay2016multi}
Attempts on motion correction and data driven image super-resolution - \cite{odille2015motion}

\appendices
\section {Comments}
Daniel: (I) Synthetic Evaluation, (Ia) Correct volume but shifted slices, (Ib) Volume with dropped out slices, (II) Coupling Parameter Sensitivity Analysis

\end{comment}

\section{Methodology}
In the next section, we briefly summarise the state-of-the-art methodology for image segmentation (SEG) and super-resolution (SR), which is based on convolutional neural networks (CNNs). We then present a novel methodology that extends these CNN models with a global training objective to constrain the output space by imposing anatomical shape priors. For this, we propose a new regularisation network that is based on the T-L architecture which was used in computer graphics \cite{girdhar2016learning} to $3$D render objects from natural images. 

\subsection{Medical Image Segmentation with CNN Models}
\label{sec:segmentationmodel}

Let $\boldsymbol{y}_s = \{y_i\}_{i \in \mathcal{S}} $ be an image of class labels representing different tissue types with $y_i \in \mathcal{L} = \{ 1,2,\hdots C \}$. Furthermore let $\boldsymbol{x} = \{  x_i \in \mathbb{R} , i \in \mathcal{S}  \}$ be the observed intensity image. The aim of image segmentation is to estimate $\boldsymbol{y}_s$ having observed $\boldsymbol{x}$. In CNN based segmentation models \cite{kamnitsas2017efficient,long2015fully,ronneberger2015u}, this task is performed by learning a discriminative function that models the underlying conditional probability distribution $P(\boldsymbol{y}_s | \boldsymbol{x})$. 

The estimation of class densities $P(\boldsymbol{y}_s | \boldsymbol{x})$ consists in assigning to each $x_i$ the probability of belonging to each of the $C$ classes, yielding $C$ sets of class feature maps $f_{c}$ that are extracted through learnt non-linear functions. The final decision for class labels is then made by applying softmax to the extracted class feature maps, in the case of cross-entropy $L_{x} = - \sum_{c=1} ^ C \, \sum_{i \in \mathcal{S}} \, \log\left(\frac{e^{f_{(c,i)}}}{ \sum_j e^{f_{(j,i)}} }\right)$ these feature maps correspond to log likelihood values.

As in the U-Net \cite{ronneberger2015u} and DeepMedic \cite{kamnitsas2017efficient} models, we learn the mapping between intensities and labels $\phi (\boldsymbol{x}) : \mathcal{X} \rightarrow \mathcal{L}$ by optimising the average cross-entropy loss of each class $L_{x} = \sum_{c=1}^C \, L_{(x,c)}$ using stochastic gradient descent. As shown in Fig. \ref{fig:seg_sr_networks}, the mapping function $\phi$ is computed by passing the input image through a series of convolution layers and rectified linear units across different image scales to enlarge the model's receptive field. The presented model is composed of two parts: feature extraction (analysis) similar to a VGG-Net \cite{simonyan2014very} and reconstruction (synthesis) as in the case of a 3D U-Net \cite{ronneberger2015u}. However, in contrast to existing approaches, we aim for sub-pixel segmentation accuracy by training up-sampling layers with high-resolution ground-truth maps. This enables $3$D analysis of the underlying anatomy in case of thick slice 2D image stack acquisitions such as cine cardiac MR imaging. In this way, it is possible to perform analysis on the high-resolution image grid without any preceding upsampling operation with a SR model \cite{oktay2016multi}. 

Similar segmentation frameworks (cf. \cite{litjens2017survey}) have been studied in medical imaging. However, in most of the existing methods, the models are supervised purely through a local loss function at pixel level (\emph{e.g.} cross-entropy, Dice) without exploiting the global dependencies and structure in the output space. For this reason, the global description of predictions is usually not adhering to shape, label, or atlas priors. In contrast to this we propose a model that can incorporate the aforementioned priors in segmentation models. The proposed framework relies on autoencoder and T-L network models to obtain a non-linear compact representation of the underlying anatomy, which are used as priors in segmentation.

\subsection{Convolutional Autoencoder Model and ACNN-Seg}
\label{sec:autoencoder}

\begin{figure}[t]
	\centering
	\includegraphics[width=.40\textwidth]{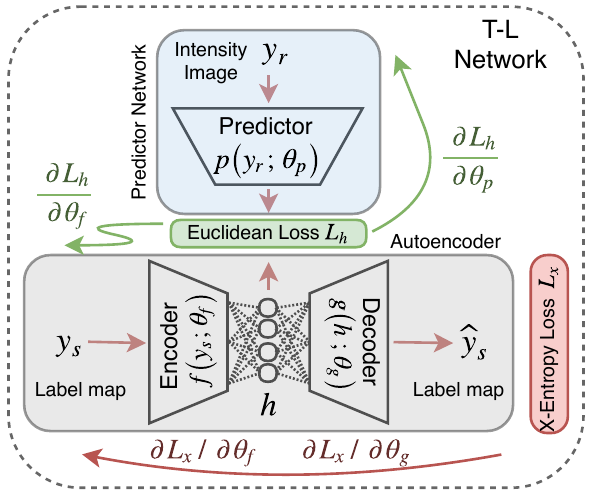}
	\caption{Block diagram of the stacked convolutional autoencoder (AE) network (in grey), which is trained with segmentation labels. The AE model is coupled with a predictor network (in blue) to obtain a compact non-linear representation that can be extracted from both intensity and segmentation images. The whole model is named as T-L network.}
	\label{fig:ae_tl_network}
\end{figure}

An autoencoder (AE) \cite{vincent2010stacked} is a neural network that aims to learn an intermediate representation from which the original input can be reconstructed. Internally, it has a hidden layer $\boldsymbol{h}$ whose activations represent the input image, often referred as {\em codes}. To avoid the AE to directly copy its output, the AE are often designed to be undercomplete so that the size of the code is less than the input dimension as shown in Fig. \ref{fig:ae_tl_network}. Learning an AE forces the network to capture the most salient features of the training data. The learning procedure minimises a loss function {\blue $L_x (\boldsymbol{y}_s, g(f(\boldsymbol{y}_s)))$, where $L_x$ is penalising $g(f(\boldsymbol{y}_s))$ being dissimilar from $\boldsymbol{y}_s$.} The functions $g$ and $f$ are defined as the decoder and encoder components of the AE.

In the proposed method, the AE is integrated into the standard segmentation network, described in Sec. \ref{sec:segmentationmodel}, as a regularisation model to constrain class label predictions $\boldsymbol{y}$ towards anatomically meaningful and accurate outputs. The cross-entropy loss function operates on individual pixel level class predictions, which does not guarantee global consistency and plausible anatomical shapes even though the segmentation network has a receptive field larger than the size of structures to be segmented. This is due to the fact that back-propagated gradients are parametrised only by pixel-wise individual probability divergence terms and thus provide little global context.

\begin{figure}[t]
	\centering
	\includegraphics[width=.50\textwidth]{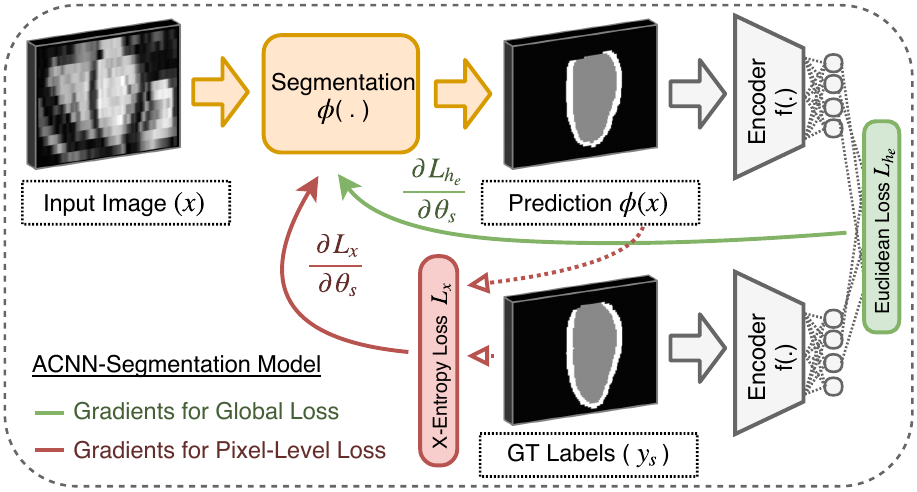}
	\includegraphics[width=.50\textwidth]{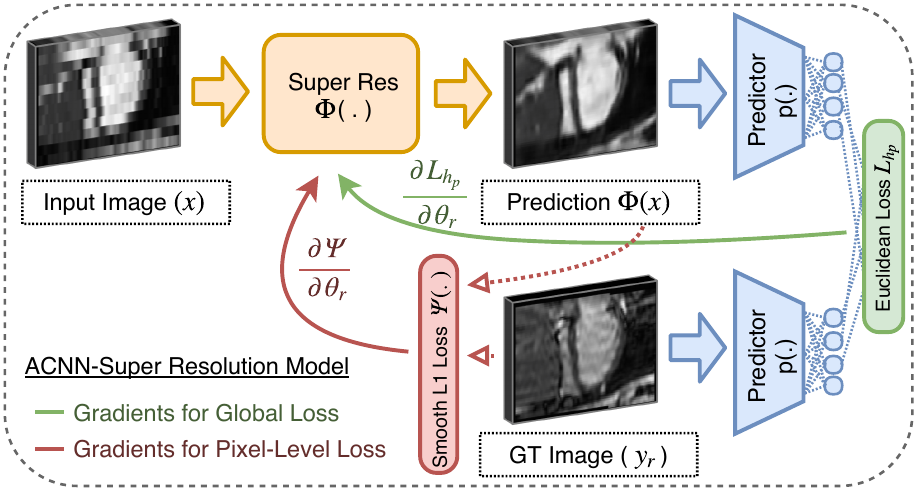}
	\caption{Training scheme of the proposed anatomically constrained convolutional neural network (ACNN) for image segmentation and super-resolution tasks. The proposed T-L network is used as a regularisation model to enforce the model predictions to follow the distribution of the learnt low dimensional representations or priors.}
	\label{fig:acnn_network}
\end{figure}

To overcome this limitation, class prediction label maps are passed through the AE to obtain a lower dimensional (\emph{e.g.} $64$ dimensions) parametrisation of the segmentation and its underlying structure \cite{sharma2016vconv}. By performing AE-based non-linear lower dimensional projections on both predictions and ground-truth labels, as shown in Fig. \ref{fig:acnn_network}, we can build our ACNN-Seg training objective function though a linear combination of cross-entropy ($L_x$), shape regularisation loss ($L_{h_e}$), and weight decay terms as follows: 
\begin{equation}
\blue
\begin{gathered}
L_{h_e} = \norm{\,f (\phi (\boldsymbol{x});\boldsymbol{\theta}_f) - f(\boldsymbol{y};\boldsymbol{\theta}_f)\,}_2^2 \\[1 mm]
\min\limits_{\boldsymbol{\theta}_s} \: \left( L_{x} \, (\phi(\boldsymbol{x};\boldsymbol{\theta}_s), \boldsymbol{y}) + \lambda_1 \cdot L_{h_e} + \frac{\lambda_2}{2} ||\boldsymbol{w}||_2^2 \right)
\end{gathered}
\end{equation}

Here $\boldsymbol{w}$ corresponds to weights of the convolution filters, and $\boldsymbol{\theta_s}$ denotes all trainable parameters of the segmentation model and only these parameters are updated during training. {\blue The coupling parameters $\lambda_1$ and $\lambda_2$ determine the weights of shape regularisation loss and weight decay terms used in the training.} In this equation, the second term $L_{h_e}$ ensures that the generated segmentations are in a similar low dimensional space (\emph{e.g.} shape manifold) as the ground-truth labels. In addition to imposing shape regularisation, this parametrisation encourages label consistency in model predictions, and reduces false-positive detections as they can influence the predicted codes in the hidden layer. The third term corresponds to weight decay to limit the number of free parameters in the model to avoid over-fitting. The proposed AE model is composed of convolutional layers and a fully connected layer in the middle as shown in Fig. \ref{fig:ae_tl_network}, which is similar to the stacked convolutional autoencoder model proposed in \cite{masci2011stacked}. The AE model details (\emph{e.g.} layer configuration, parameter choices) are provided in the supplementary material. 

\subsection{Medical Image Super-Resolution (SR) with CNNs}

Super-resolution (SR) image generation is an inverse problem where the goal is to recover spatial frequency information that is outside the spatial bandwidth of the low resolution (LR) observation $\boldsymbol{x} \in \mathbb{R}^N$ to predict a high resolution (HR) image $\boldsymbol{y}_r \in \mathbb{R}^M$ ($N \ll M$), as illustrated in the top row of Fig. \ref{fig:ClinicalMotivation}. Since the high frequency components are missing in the observation space, usually training examples are used to predict the most likely $P(\boldsymbol{y}_r | \boldsymbol{x})$ HR output. Image SR is an ill-posed problem as there are an infinite number of solutions for a given input sample but only a few would be anatomically meaningful and accurate. As for the case of image segmentation, learnt shape representations can be used to regularise image SR, constraining the model to make only anatomically meaningful predictions. 

Similar to the SR framework described in \cite{oktay2016multi}, our proposed SR model learns a mapping function $\Phi : \mathcal{X} \rightarrow \mathcal{Y}$  to estimate a high-resolution image $\hat{\boldsymbol{y}}_r = \Phi(\boldsymbol{x}\,; \boldsymbol{\theta}_r)$ where $\boldsymbol{\theta}_r$ denotes the model parameters such as convolution kernels and batch-normalisation statistics. The parameters are optimised by minimising the smooth $\ell_1$ loss, also known as Huber loss, between the ground-truth high resolution image and the corresponding prediction. The smooth $\ell_1$ norm is defined as $\Psi_{\ell_1} (k) = \{ 0.5\, k^2 \; \text{if} \; |k|<1 \, , \, |k|-0.5 \; \text{otherwise} \}$  and the SR training objective becomes $\min\limits_{\boldsymbol{\theta}_r} \, \sum_{i \in \mathcal{S}} \, \Psi_{\ell_1} \left( \Phi (\boldsymbol{x}_i \,; \boldsymbol{\theta}_r) - \boldsymbol{y}_i \right)$

In the proposed SR framework, we used the same model as shown in Fig. \ref{fig:seg_sr_networks}. It provides two main advantages over the state-of-the-art medical image SR model proposed in \cite{oktay2016multi}: (I) the network generates image features in the LR image grid rather than early upsampling of the features, which reduces memory and computation requirements significantly. As highlighted in \cite{shi2016real}, early upsampling introduces redundant computations in the HR space since no additional information is added into the model by performing transposed convolutions \cite{zeiler2010deconvolutional} at an early stage. (II) The second advantage is the use of a larger receptive field to learn the underlying anatomy, which was not the case in earlier SR methods used in medical imaging \cite{oktay2016multi} and natural image analysis \cite{shi2016real, dong2016accelerating} because these models usually operate on local patch level. Capturing large context indeed helps our model to better understand the underlying anatomy and this enables us to enforce global shape constraints. This is achieved by generating SR feature-maps in multiple scales using multi strides in the in-plane direction.

Similar to the ACNN-Seg model, it is possible to regularise SR models to synthesise anatomically more meaningful HR images. To achieve this goal, we extend the standard AE model to the T-L model which enables us to obtain shape representation codes directly from the intensity space. The idea is motivated by the recent work \cite{girdhar2016learning} on $3$D shape analysis in natural images. In the next section we will explain the training strategy and the use of the T-L model as a regulariser.

\begin{comment}
plan 

- we added the training scheme of TL learning. 

- we added why we need this architecture, basically it helps us to reach codes from images, whereas only autoencoder network requires segmentation. We do not always need to obtain segmentation to identify the codes. This is the main reason why we use this architecture. 

\end{comment}

\subsection{T-L Network Model and SR-ACNN}

{\blue Shape encoding AE models operate only on the segmentation masks and this limits its application to SR problem where the model output is an intensity image. To circumvent this problem, we extend the standard denoising AE to the T-L regularisation model by combining the AE with a predictor network (Fig. \ref{fig:ae_tl_network}) $p(\boldsymbol{x}) : \mathcal{X} \rightarrow \mathcal{H}$. The predictor can map an input image into a low dimensional non-parametric representation of the underlying anatomy (\emph{e.g.} shape and class label information), which is learnt by the AE. In other words, it enables us to learn a hidden representation space that can be reached by non-linear mappings from both image label space $\mathcal{Y}$ and image intensity space $\mathcal{X}$. In this way, SR models can be regularised as well with respect to learnt anatomical priors.}

This network architecture is useful in image analysis applications for two main reasons: (I) It enables us to build a regularisation network that could be used in applications different than image segmentation such as image SR. We propose to use this new regularisation network at training time of SR to enforce the models to learn global information about the images besides the standard pixel-wise ($\ell_1$ distance) image reconstruction loss. In this way, the regressor SR model is guided by the additional segmentation information, and it becomes robust against imaging artefacts and missing information. (II) The second important feature of the T-L model is the generalisation of the learnt representations. Joint training of the AE and predictor enables us to learn representations that could be extracted from both intensity and label space. The learnt codes will encode the variations that could be interpreted from both manual annotations and intensity images. Since a perfect mapping between the intensity and label spaces is practically not achievable, the T-L learnt codes are expected to be more representative due to the inclusion of additional information.

The T-L model is trained in two stages: In the first stage, the AE is trained separately {\blue with ground-truth segmentation masks and cross-entropy loss $L_x$}. Later, the predictor model is trained to match {\blue the learnt latent space $\boldsymbol{h}$ by minimising the Euclidean distance $L_h$ between the codes predicted by the AE and predictor as shown in Fig. \ref{fig:ae_tl_network}}. Once the loss functions for both the AE and the predictor converge, the two models are trained jointly in the second stage. {\blue The encoder $f$ is updated using two separate back-propagated gradients $(\frac{\partial L_x}{\partial \theta_f}\, , \frac{\partial L_h}{\partial \theta_f})$ and the two loss functions are scaled to match their range. The first gradient encourages the encoder to generate codes that could be easily extracted by the predictor while the second gradient} making sure that a good segmentation-reconstruction can be obtained at the output of the decoder. {\blue Training details are further discussed in Section \ref{sec:trainingdetails}.} It is important to note that the T-L regulariser model is used only at training time but not during inference; in other words, the fully convolutional (FCN) segmentation and super-resolution models can still be used for applications using different image sizes. In this paper, the proposed SR model is referred to as ACNN-SR and its training scheme is shown in the bottom part of Fig. \ref{fig:acnn_network}.

\begin{equation}
\blue
\label{eqn:acnnsr}
\begin{gathered}
L_{h_p} = \norm{\, p\,(\Phi (\boldsymbol{x}); \boldsymbol{\theta}_p) - p\,(\boldsymbol{y}_r;\boldsymbol{\theta}_p)\,}_2^2 \\[1 mm]
\min\limits_{\boldsymbol{\theta}_r} \, \left( \Psi_{\ell_1} \left( \Phi (\boldsymbol{x}\,; \boldsymbol{\theta}_r ) - \boldsymbol{y}_r \right) + \lambda_1 \cdot L_{h_p} + \frac{\lambda_2}{2} ||\boldsymbol{w}||_2^2  \right)
\end{gathered}
\end{equation}

 The training objective shown above is composed of weight decay, pixel-wise and global loss terms. {\blue Here $\lambda_1$ and $\lambda_2$ determine the weight of shape priors and weight decay terms} while the smooth $\ell_1$ norm loss function $\Psi$ quantifies the reconstruction error. The global loss $L_{h_p}$ is defined as the Euclidean distance between the codes generated from the synthesised and ground-truth HR images. The T-L model is used only in the network training phase as a regularisation term, similar to VGG features \cite{simonyan2014very} that were used for representing a perceptual loss function \cite{johnson2016perceptual}. However, we are not interested in expanding the output space to a larger feature-map space, but instead obtain a compact representation of the underlying anatomy. 

\subsection{Learnt Hidden Representations}
\label{sec:learntrepresentations}
\begin{figure}[t]
 	\centering
 	\includegraphics[width=.49\textwidth]{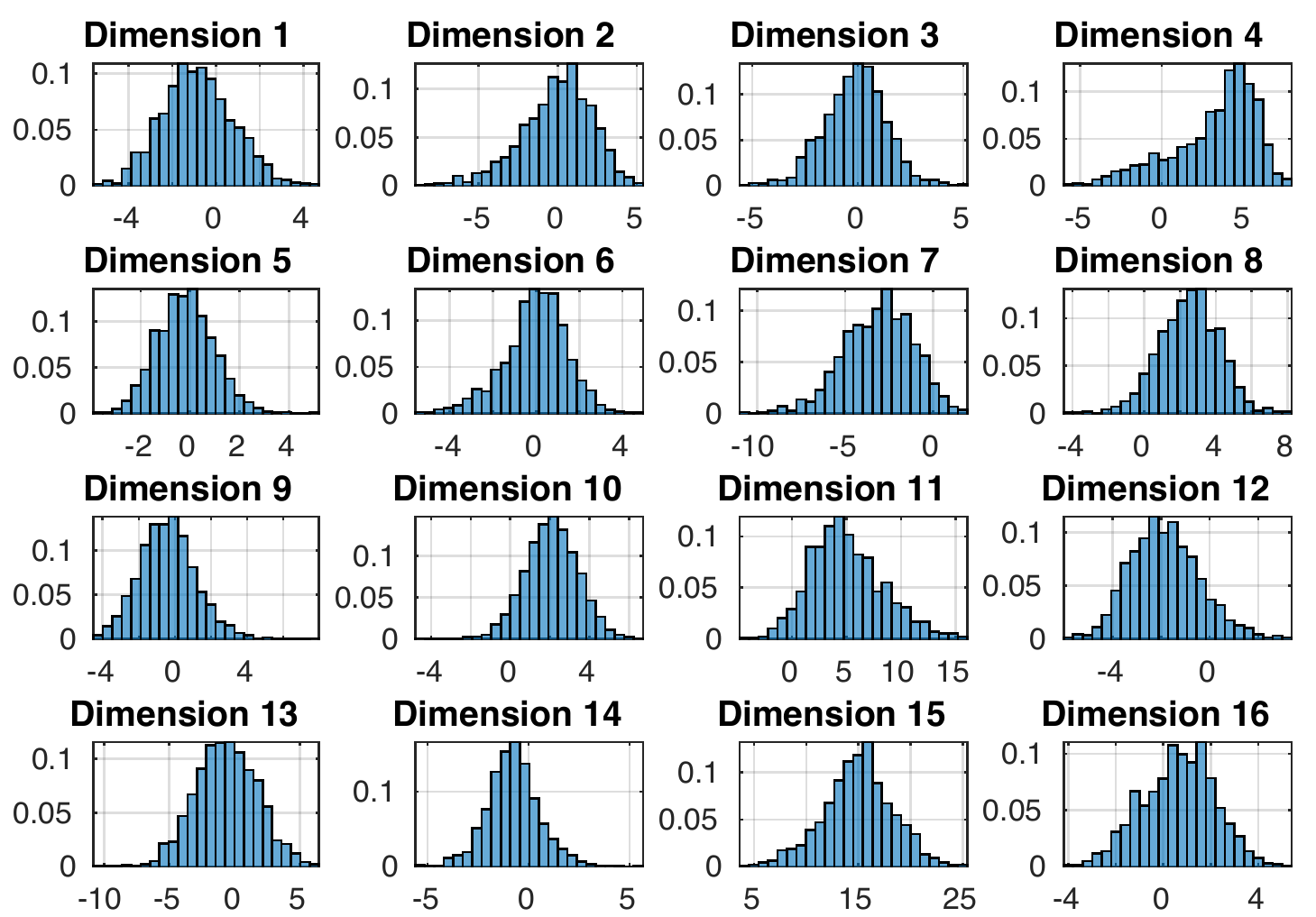}
 	\caption{Histogram of the learnt low-dimensional latent representations (randomly selected $16$ components are shown). The codes in general follow a smooth and normal distribution which is important for the training of ACNN models.)}
 	\label{fig:code_visualisation}
 \end{figure}

 The learnt low dimensional representation $\boldsymbol{h}$ is used to constrain NN models. Low dimensional encoding enables us to train models with global characteristics but also yields better generalisation power for the underlying anatomy as shown in earlier work \cite{torralba2008small}. However, since we update our segmentation and SR model parameters with the gradients back-propagated from the global loss layer using the Euclidean distance of these representations, it is essential to analyse the distribution of the extracted codes. In Fig. \ref{fig:code_visualisation}, due to space limitations, we show the histogram of $16$ randomly chosen codes (out of $64$) of a T-L model trained with cardiac MR segmentations. Note that each histogram is constructed using the corresponding code for every sample in the full dataset. It is observed that the learnt latent representations in general follow a normal distribution and they are not separated in multi-clusters (\emph{e.g.} mixture of Gaussians). A smooth distribution of the codes ensures better supervision for the main NN model (SR, Seg) since the global gradients are back-propagated by computing the Euclidean distance between the obtained distributions. 

This observation can be explained by the fact that the proposed T-L network is trained with small Gaussian input noise as in the case of denoising autoencoders. In \cite{alain2014regularized}, Alain and Bengio showed that the denoising reconstruction error is equivalent to contractive penalty, which forces the feature extraction (encoder) function $f$ resist perturbations of the input and contracts these input samples to similar low dimensional codes. The penalty is defined as $\Omega (\boldsymbol{h}) = \lambda \norm{ \frac{\partial f(\boldsymbol{x})}{\partial \boldsymbol{x}} }_F^2$, where $F$ denotes the Frobenius norm (sum of squared elements), and $\boldsymbol{h} = f(\boldsymbol{x})$ represents the codes. The given penalty function promotes the network to learn the underlying low-dimensional data manifold and capture its local smooth structure. In addition to the smoothness of the latent distributions, the extracted codes are expected to be correlated since the decoder merges some of the codes along the three spatial dimensions to construct input feature maps for the transposed convolutions, but this characteristic is not a limitation in our study. 

\begin{table*}[t]
\parbox{\textwidth}{
\centering
\captionof{table}{Stacks of 2D cardiac MR images ($200$) are segmented into LV endocardium and myocardium, and the segmentation accuracy is evaluated in terms of Dice metric and surface to surface distances. The ground-truth labels are obtained from high resolution 3D images acquired from same subjects, which do not contain motion and blocky artefacts. The proposed approach (ACNN-Seg) is compared against state-of-the-art slice by slice segmentation (2D-FCN \cite{tran2016fully}) method, {\blue 3D-UNet model \cite{cicek20163d}, cascaded 3D-UNet and convolutional AE model (AE-Seg) \cite{ravishankarlearning}}, proposed sub-pixel segmentation model (3D-Seg) and the same model with motion augmentation used in training (3D-Seg-MAug).}
\begin{tabular}{@{\extracolsep{5pt}}lccccccc@{}}
\phantom{a} & \multicolumn{3}{c}{Endocardium} & \multicolumn{3}{c}{Myocardium} & Capacity \\
\cmidrule{2-4} \cmidrule{5-7} \cmidrule{8-8}
& Mean       & Hausdorff  & Dice       & Mean       & Hausdorff  & Dice       & \# Trainable \\
& Dist. (mm) & Dist. (mm) & Score (\%) & Dist. (mm) & Dist. (mm) & Score (\%) & Parameters\\
\midrule

2D-FCN \cite{tran2016fully} & 2.07$\pm$0.61 & 11.37$\pm$7.15 & .908$\pm$.021 & 1.58$\pm$0.44 & 9.19$\pm$7.22 & .727$\pm$.046 & $1.39\times10^6$ \\ 
3D-Seg & 1.77$\pm$0.84 & 10.28$\pm$8.25 & .923$\pm$.019 & 1.48$\pm$0.51 & 10.15$\pm$10.58 & .773$\pm$.038 & $1.60 \times 10^6$ \\
{\blue 3D-UNet \cite{cicek20163d}} & {\blue 1.66$\pm$0.74} & {\blue 9.94$\pm$9.22} & {\blue .923$\pm$.019} & {\blue 1.45$\pm$0.47} & {\blue 9.81$\pm$11.77} & {\blue .764$\pm$.045} & {\blue $1.64 \times 10^6$}\\

{\blue AE-Seg \cite{ravishankarlearning}} & {\blue 1.75$\pm$0.58} & {\blue{8.42$\pm$3.64}} & {\blue .926$\pm$.019} & {\blue 1.51$\pm$0.29} & {\blue {8.52$\pm$2.72}} & {\blue .779$\pm$.033} & {\blue $1.68\times10^6$}\\

3D-Seg-MAug          & 1.59$\pm$0.74 & 8.52$\pm$8.13  & .928$\pm$.019 & 1.37$\pm$0.41 & 9.41$\pm$9.17   & .785$\pm$.041 & $1.60 \times 10^6$ \\

{\blue AE-Seg-M} & {\blue 1.59$\pm$0.48} & \textbf{{\blue{7.52$\pm$3.78}}} & {\blue .927$\pm$.017} & {\blue 1.32$\pm$0.26} & \textbf{{\blue {7.12$\pm$2.79}}} & {\blue .791$\pm$.036} & {\blue $1.91\times10^6$}\\

\textbf{ACNN-Seg} & \textbf{1.37}$\pm$\textbf{0.42} & 7.89$\pm$3.83  & \textbf{.939}$\pm$\textbf{.017} & \textbf{1.14}$\pm$\textbf{0.22} & 7.31$\pm$3.59   & \textbf{.811}$\pm$\textbf{.027} & $1.60 \times 10^6$\\
{\blue p-values}  & {\blue $p \ll 0.001$} & {\blue $p \approx 0.890$} & {\blue $p \ll 0.001$} & {\blue $p \ll 0.001$} & {\blue $p \approx 0.071$} & {\blue $p \ll 0.001$} & {\blue -} \\
\bottomrule
\end{tabular}
 \label{tab:CMRSEG_results}
}
\end{table*}

\section{Applications and Experiments}
In this section, we present three different applications of the proposed ACNN model: 3D-US and cardiac MR image segmentation, as well as cardiac MR image SR. The experiments focus on demonstrating the importance of shape and label priors for image analysis. Additionally, we analyse the salient information stored in the learnt hidden representations and correlate them with clinical indices, showing their potential use as biomarkers for pathology classification. The next subsection describes the clinical datasets used in our experiments. 

\subsection{Clinical Datasets}
	\subsubsection{UK Digital Heart Project Dataset} \label{sec:UKDigital} This dataset \footnote{https://digital-heart.org/} is composed of $1200$ pairs of cine 2D stack short-axis (SAX) and cine 3D high resolution (HR) cardiac MR images. Each image pair is acquired from a healthy subject using a standard imaging protocol \cite{bai2015bi,de2014population}. In more detail, the 2D stacks are acquired in different breath-holds and therefore may contain motion artefacts. Similarly, 3D imaging is not always feasible in the standard clinical setting due to the requirements for long image acquisition. The voxel resolution of the images are fixed to $1.25$x$1.25$x$10.00$ mm and $1.25$x$1.25$x$2.00$ mm for 2D stack low resolution (LR) and HR images respectively. Dense segmentation annotations for HR images are obtained by manually correcting initial segmentations generated with a semi-automatic multi-atlas segmentation method \cite{bai2013probabilistic}, and all the annotations are performed on the HR images to minimise errors introduced due to LR in through plane direction. {\blue Since the ground-truth information is obtained from the HR motion-free images, the experimental results are expected to reflect the performance of the method with respect to an appropriate reference.} The annotations consist of pixel-wise labelling of endocardium and myocardium classes. Additionally, the residual spatial misalignment between the 2D LR stacks and HR volumes is corrected using a rigid transformation estimated by an intensity based image registration algorithm.  

	\subsubsection{CETUS'14 Challenge Dataset} \label{sec:CETUS} CETUS'14 segmentation challenge \cite{bernard2016standardized} is a publicly available platform \footnote{https://www.creatis.insa-lyon.fr/Challenge/CETUS/index.html} to benchmark cardiac 3D ultrasound (US) left-ventricle (LV) segmentation methods. The challenge dataset is composed of 3D+time US image sequences acquired from $15$ healthy subjects and $30$ patients diagnosed with myocardial infarction or dilated cardiomyopathy. The images were acquired from apical windows and LV chamber was the main focus of analysis. Resolution of the images was fixed to $1$ mm isotropic voxel size through linear interpolation. The associated manual contours of the LV boundary were drawn by three different expert cardiologists, and the annotations were performed only on the frames corresponding to end-diastole (ED) and end-systole (ES) phases. Method evaluation is performed in a blinded fashion on the testing set ($30$ out of $45$) using the MIDAS web platform.
	
	\subsubsection{ACDC MICCAI'17 Challenge Dataset} \label{sec:ACDC} The aim of the ACDC'17 challenge \footnote{https://www.creatis.insa-lyon.fr/Challenge/acdc/} is to compare the performance of automatic methods for the classification of MR image examinations in terms of healthy and pathological cases: infarction, dilated cardiomyopathy, and hypertrophic cardiomyopathy. The publicly available dataset consists of $20$ (per class) cine stacks of 2D MR image sequences which are annotated at ED and ES phases by a clinical expert.  In the experiments, latent representations ({\em codes}) extracted with the proposed T-L network are used to classify these images. 
	
{\blue \subsection{Training Details of the Proposed Model}
\label{sec:trainingdetails}
In this section, we discuss the details of data augmentation used in training, and also the optimisation scheme of the T-L model training. To improve the model's generalisation capability, the input training samples are artificially augmented using affine transformations, which is used in both the segmentation and T-L models.  For the SR models, on the other hand, respiratory motion artefacts between the adjacent slices are simulated via in-plane rigid transformations that are defined for each slice independently. The corresponding ground-truth HR images are not spatially transformed; in this way, the models learn to output anatomically correct results when the input slices are motion corrupted. Additionally, additive Gaussian noise is applied to input intensity images to make the segmentation and super-resolution models more robust against image noise. For the AE, the tissue class labels are randomly swapped with the probability of $0.1$ to encourage the model to map slightly different segmentation masks to neighbouring points in the lower dimensional latent space. It ensures the smoothness of the learnt low-dimensional manifold space as explained in Section \ref{sec:learntrepresentations}.

In the joint training of the T-L network, parameters of the encoder model ($f$) are updated by the gradients originating from both the cross-entropy loss ($L_x$) and Euclidean distance terms ($L_h$). Instead of applying these two gradient descent updates sequentially in an iterative fashion, we perform a joint update training scheme and experimentally observed better convergence.}

\begin{table*}[t]
	\parbox{\textwidth}{
		\centering
		\captionof{table}{3D-US cardiac image sequences (in total $30$) are segmented into LV cavity and background. Segmentation accuracy is evaluated in terms of Dice score (DSC), surface-to-surface distances. The consistency of delineations on both ED and ES phases are measured in terms computed ejection fraction (EF) values. The proposed ACNN-Seg method is compared against state-of-the art deformable shape fitting \cite{barbosa2013fast} and fully-convolutional 3D segmentation \cite{chen2016iterative} methods. }
		
		\begin{tabular}{@{\extracolsep{3pt}}lccccccc@{}}
			\phantom{a} & \multicolumn{3}{c}{End Diastole (ED)} &  \multicolumn{3}{c}{End Systole (ES)} \\
			\cmidrule{2-4} \cmidrule{5-7} 
			& BEAS \cite{barbosa2013fast} & FCN \cite{chen2016iterative} & ACNN-Seg & BEAS \cite{barbosa2013fast} & FCN \cite{chen2016iterative} & ACNN-Seg & {\blue p-values}\\ \midrule
			
			Mean Dist (mm)&   2.26$\pm$0.73  & 1.98$\pm$1.03  &  \textbf{1.89}$\pm$\textbf{0.51}  & 2.43$\pm$0.91 & 2.83$\pm$1.89 & \textbf{2.09}$\pm$\textbf{0.77} & {\blue $p < 0.01$}  \\ 
			
			HD Dist (mm)  &   8.10$\pm$2.66 & 11.94$\pm$9.46 &  \textbf{6.96}$\pm$\textbf{1.75}  & 8.13$\pm$3.08 & 12.45$\pm$10.69 &  \textbf{7.75}$\pm$\textbf{2.65} & {\blue $p < 0.001$}  \\
			
			DSC (\%)      &   .894$\pm$.041  & .906$\pm$.026 & \textbf{.912}$\pm$\textbf{.023}  & .856$\pm$.057 & .\textbf{872}$\pm$\textbf{.050} & \textbf{.873}$\pm$\textbf{.051} & {\blue $p \approx 0.05$} \\
			
			EF (Corr)     &   0.889          & 0.885          &  \textbf{0.913}   & -  & - & - & - \\
			
			EF (Bias+LOA) (ml) &   -6.78$\pm$27.71  & 2.74$\pm$12.01  &   \textbf{1.78}$\pm$\textbf{10.09}  & -  & - & - & - \\
			\bottomrule
		\end{tabular}
		\label{tab:3DUS_results}
	}
\end{table*}

\subsection{Cardiac Cine-MR Image Segmentation} 
In this experiment, NN models are used to segment cardiac cine MR images in the dataset described in Sec. \ref{sec:UKDigital}. As an input to the models, only the 2D stack LR images are used, which is a commonly used acquisition protocol for cardiac imaging, and the segmentation is performed only on the ED phase of the sequences. The corresponding ground-truth label maps, however, are projected from the HR image space, which are annotated in the HR image grid. The dataset ($1200$ LR images \& HR labels) is randomly partitioned into three subsets: training ($900$), validation ($100$), and testing ($200$). All the images are linearly intensity normalised and cropped based on the automatically detected six anatomical landmark locations \cite{oktay2017stratified}. %Other segmentation approaches employed simpler techniques (\emph{e.g.} Hough Transform) to localise the heart in MR images. 

	\begin{figure}[t!]
		\centering
		\includegraphics[width=.45\textwidth]{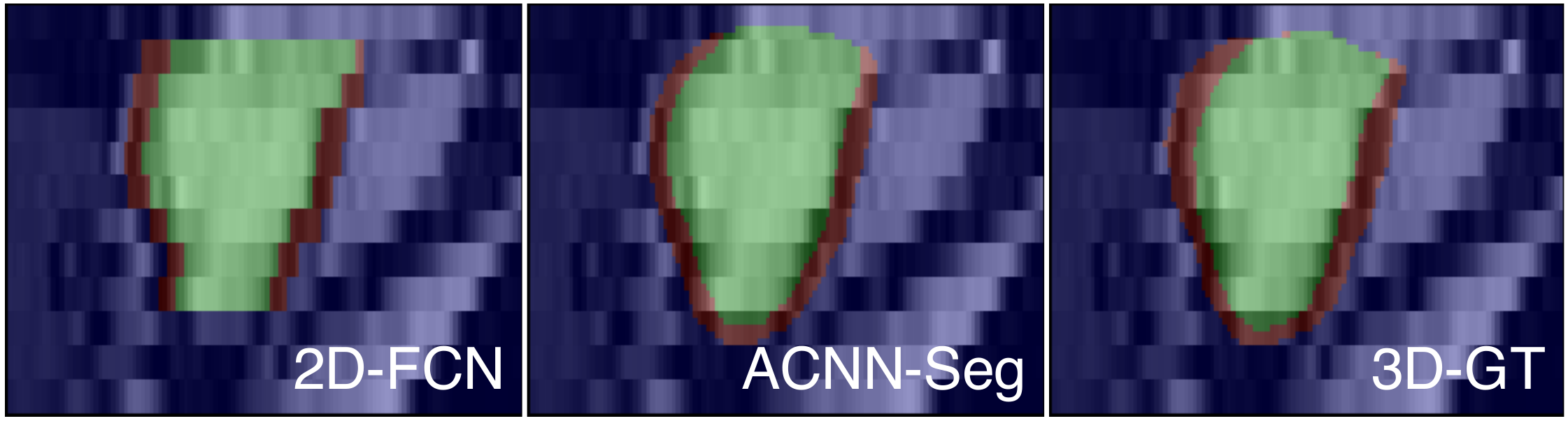}
		\includegraphics[width=.45\textwidth]{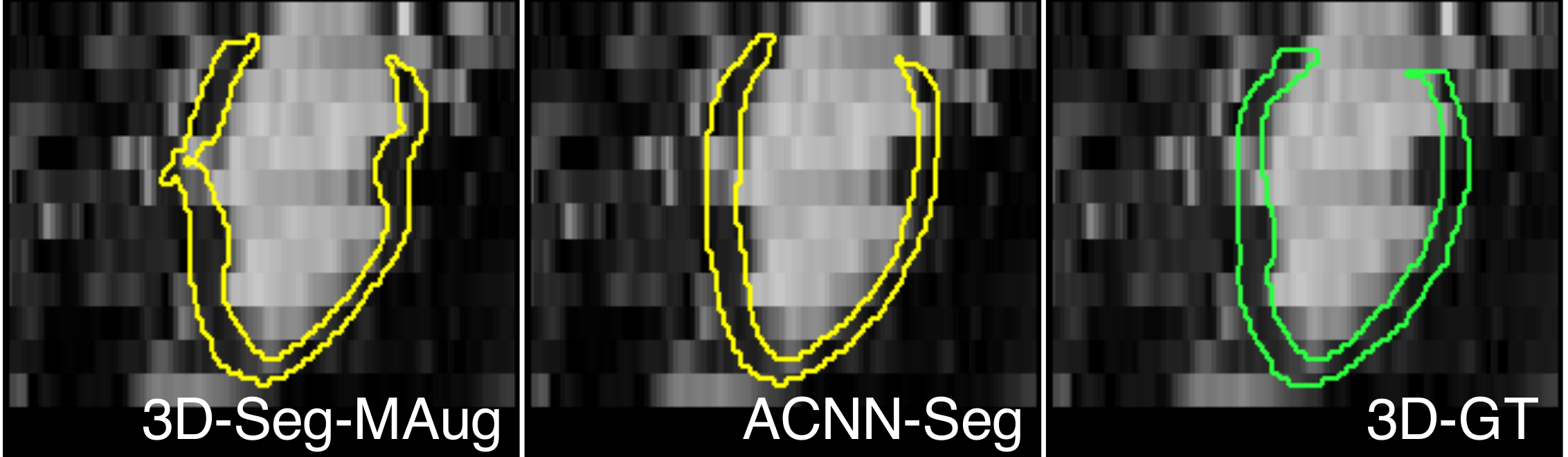}
		\caption{Segmentation results on two different 2D stack cardiac MR images. The proposed ACNN model is insensitive to slice misalignments as it is anatomically constrained and it makes less errors in basal and apical slices compared to the 2D-FCN approach. The results generated from low resolution image is better correlated with the HR ground-truth annotations (green).}
		\label{fig:results3DMRUSSEG}
	\end{figure}	
	
The proposed ACNN-Seg method is compared against: the current state-of-the-art cine MR 2D slice by slice segmentation method (2D-FCN) \cite{tran2016fully}, {\blue 3D-UNet model \cite{cicek20163d}, cascaded 3D-UNet and convolutional AE model (AE-Seg) \cite{ravishankarlearning}}, sub-pixel 3D-CNN segmentation model (3D-Seg) proposed in Sec. \ref{sec:segmentationmodel}, and the same model trained with various types of motion augmentation (3D-Seg-MAug). {\blue As the models have a different layout, the number of trainable parameters (pars) used in each model is kept fixed to avoid any bias. For the cascaded AE-Seg model, however, additional convolutional kernels are used in the AE as suggested in \cite{ravishankarlearning}. To observe the influence of the AE model's capacity on the AE-Seg model's performance, we performed experiments using different number of AE pars, and the largest capacity case is denoted by AE-Seg-M. 
	
The results of the experiments are provided in Table \ref{tab:CMRSEG_results} together with the capacity of each model. Statistical significance of the results is verified by performing the Wilcoxon signed-rank test between the top two performing methods for each evaluation metric.} Based on these results we can draw three main conclusions: (I) Slice by slice analysis \cite{avendi2016combined,tran2016fully} significantly under-performs compared to the proposed sub-pixel and ACNN-Seg segmentation methods. In particular, the dice score metrics are observed to be lower since 2D analysis can yield poor performance in basal and apical parts of the heart as shown in Fig. \ref{fig:results3DMRUSSEG}. Previous slice by slice segmentation approaches validated their methods on LR annotations; however, we see that the produced label maps are far off from the true underlying ventricular geometry and it can be a limiting factor for the analysis of ventricle morphology.  Similar results were obtained in clinical studies \cite{de2014population}, which however required HR image acquisition techniques.  (II) The results also show that introduction of shape priors in segmentation models can be useful to tackle false-positive detections and motion-artefacts. As can be seen in the bottom row of Fig. \ref{fig:results3DMRUSSEG}, without the learnt shape priors, label map predictions are more prone to imaging artefacts. Indeed, it is the main reason why we observe such a large difference in terms of Hausdorff distance. For endocardium labels, on the other hand, the difference in dice score metric is observed to be less due to the larger size of the LV blood pool compared to the myocardium. 

{\blue Lastly (III), we observe a performance difference between the cascaded AE based segmentation (AE-Seg \cite{ravishankarlearning}) and the proposed ACNN-Seg models: the segmentations generated with the former model are strongly regularised due to the second stage AE. It results in reduced Hausdorff distance with marginal statistical significance, but the model overlooks fine details of the myocardium surface since the segmentations are generated only from the coarse level feature-maps. More importantly, cascaded approaches add additional computational complexity due to the increased number of filters, which could be redundant given that the standard segmentation model is able to capture shape properties of the organs as long as it has a large receptive field and is optimised with shape constraints. In other words, shape constraints can be learnt and utilised in standard segmentation models, as shown in ACNN-Seg, without a need for additional model parameters and computational complexity. We also analysed the performance change in AE-Seg with respect to the number of parameters, which shows that the small capacity AE-Seg model ($8\times10^4$ pars) is not suitable for cardiac image segmentation as the second stage in the cascaded model does not improve the performance significantly. }

We performed additional segmentation experiments using only the T-L network. In detail, the input LR image is passed first through the predictor network and then the extracted codes are fed to the decoder network shown in Fig. \ref{fig:ae_tl_network}. Label map predictions are collected at the output of the decoder and they are compared with the same ground-truth annotations described previously, which was similar to the AE based segmentation method proposed in \cite{avendi2016combined,avendi2017automatic}. We observed that reconstruction of label-maps from low dimensional representations was limited since the ventricle boundaries were not delineated properly but rather a rough segmentation was generated (DSC: $.734$). We believe that this is probably the main reason why the authors of \cite{avendi2016combined} proposed the use of a separate deformable model at the output of a NN. Nevertheless, the proposed ACNN-Seg does not require an additional post-processing step. 
 
\subsection{Cardiac 3D Ultrasound Image Segmentation}
In the second experiment, the proposed model is evaluated on 3D cardiac ultrasound data which is described in Sec \ref{sec:CETUS}. Segmentation models are used to delineate endocardial boundaries and the segmentations obtained on ED and ES frames are later used to measure volumetric indices such as ejection fraction (EF). The models are compared also in terms of surface to surface distance errors of their corresponding endocardium segmentations. As a baseline CNN method, we utilised the fully convolutional network model suggested by \cite{chen2016iterative} for multi-view 3D-US image segmentation problem. It is also observed to be more memory efficient compared to the standard 3D-UNet architecture \cite{cicek20163d}. Additionally, we compare our proposed model against the CETUS'14 challenge winner approach (BEAS) \cite{barbosa2013fast} that utilised deformable models to segment the left ventricular cavity. The challenge results can be found in \cite{bernard2016standardized}.
 \begin{figure}
 	\centering
 	\includegraphics[width=.50\textwidth]{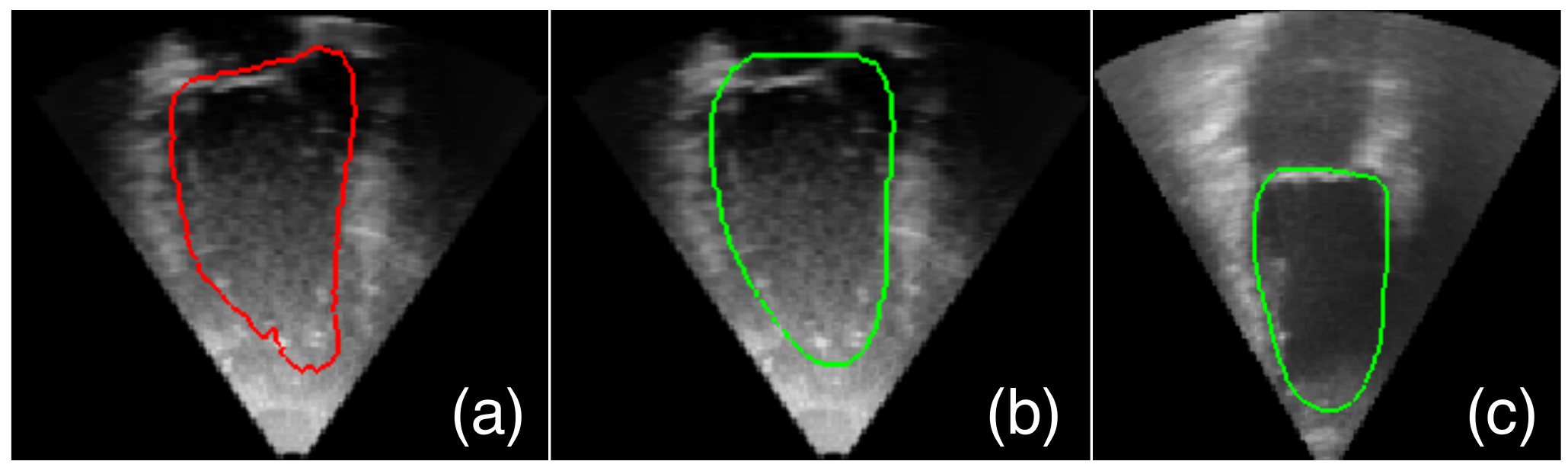}
 	\caption{(a) Cavity noise limits accurate delineation of the LV cavity in apical areas. (b) The segmentation model can be guided through learnt shape priors to output anatomically correct delineations. (c) Similarly, it can make accurate predictions even when the ventricle boundaries are occluded.}
 	\label{fig:results3DUSSEG}
 \end{figure}
The experimental results, given in Table \ref{tab:3DUS_results}, show that neural network models outperforms previous state-of-the-art approaches in this public benchmark dataset although the training data size was limited to 15 image sequences. The experimental results were evaluated in a blinded fashion by uploading the generated segmentations from separate 30 sequences into the CETUS web platform. The main contribution of ACNN model over the standard FCN approaches is the improved shape delineation of the LV, as it can be seen in terms of the distance error metrics. In particular, Hausdorff distances were reduced significantly as global regularisation reduces the amount of spurious false positive predictions and enforces abnormal LV segmentation shapes to fit into the learnt representation model. This situation is illustrated in Fig. \ref{fig:results3DUSSEG}. Similarly, we observed an improvement in terms of normalised Dice score, which was quantitatively not significant due to large volumetric size of the LV cavity. Lastly, we compared the extracted ejection fraction results to understand both the accuracy of segmentations and also the consistency of these predictions on both ED and ES phases. It is observed that the ACNN approach ensures better consistency between frames although none of the methods have used temporal information. 

The reported results could be further improved by segmenting both ED and ES frames simultaneously or by extracting the temporal content from the sequences. For instance, propagation of ED masks to ES frames through optical flow has been shown to be a promising way to achieve this goal. However, this study mainly focuses on demonstrating the advantages of using priors in neural network models, and achieving the best possible segmentation accuracy was not our main focus. 

\subsection{Cardiac MR Image Enhancement}
The proposed ACNN model is also applied to the image SR problem and compared against the state-of-the-art CNN model used in medical imaging \cite{oktay2016multi}. The cardiac MR dataset, described in Sec. \ref{sec:UKDigital}, was split into two disjoint subsets: training ($1000$) and testing ($200$). At testing time, we evaluated our model with both LR-HR clinical image data. In training, however, LR images are synthetically generated from clinical HR data using the MR acquisition model discussed in \cite{greenspan2009super}. More details about the acquisition model can be found in \cite{oktay2016multi}. 

The quality of the upsampled images is evaluated in terms of SSIM metric \cite{wang2004image} between the clinical HR image data and reconstructed HR images. SSIM measure assesses the correlation of local structures and is less sensitive to image noise than PSNR which is not used in our experiments since small misalignments between LR-HR image pairs could introduce large errors in the evaluation due to pixel by pixel comparisons. More importantly, intensity statistics of the images are observed to be different for this reason PSNR measurements would not be accurate. In addition to the SSIM metric, we used the mean opinion score (MOS) testing \cite{ledig2016photo} to quantify the quality and similarity of the synthesised and real HR cardiac images. Two expert cardiologists were asked to rate the upsampled images from 1 (very poor) to 5 (excellent) based on the accuracy of the reconstructed LV boundary and geometry. To serve as a reference, the corresponding clinical LR and HR images are displayed together with the upsampled images that are anonymised for a fair comparison. 

\begin{table}
 	\parbox{0.49\textwidth}{
 		\captionof{table}{Average inference time (Inf-T) of the SR models per input LR image (120x120x12) using a GPU (GTX-1080). ACNN-SR and SR-CNN \cite{oktay2016multi} models are given the same number of filters and capacity. MOS \cite{ledig2016photo} results, received from the clinicians (R1 and R2), are reported separately.} 
 		\begin{tabular}{@{\extracolsep{0pt}}lcccc@{}}
 			                & SSIM \cite{wang2004image} & MOS-R1         & MOS-R2              & Inf-T \\ \midrule
 			Linear                       &   .777$\pm$.043   & 2.71$\pm$0.82  & 2.60$\pm$.91  &  -  \\
	 		B-Spline                     &   .779$\pm$.053   & 2.77$\pm$0.89  & 2.64$\pm$.84 &  -  \\ 
 			SR-CNN \cite{oktay2016multi}  &   .783$\pm$.046   & 3.59$\pm$1.05  & 3.85$\pm$.70&  .29 s \\
 			{\blue 3D-UNet \cite{cicek20163d}}  & {\blue .784$\pm$.045} & {\blue 3.55$\pm$0.92} & {\blue 3.99$\pm$.71} & {\blue .07 s} \\
 			ACNN-SR                      &   \textbf{.796}$\pm$\textbf{.041}  & \textbf{4.36}$\pm$\textbf{0.62}  & \textbf{4.25}$\pm$\textbf{.68}&  \textbf{.06} s\\
 			{\blue p-values} & {\blue $p \ll 0.001$} & {\blue $p < 0.001$} & {\blue $p < 0.01$} & {\blue -} \\
 			\bottomrule
 		\end{tabular}
 		\label{tab:resultsMRSR}
 	}
\end{table}

\begin{figure}[t!]
	\centering
	\includegraphics[width=.45\textwidth]{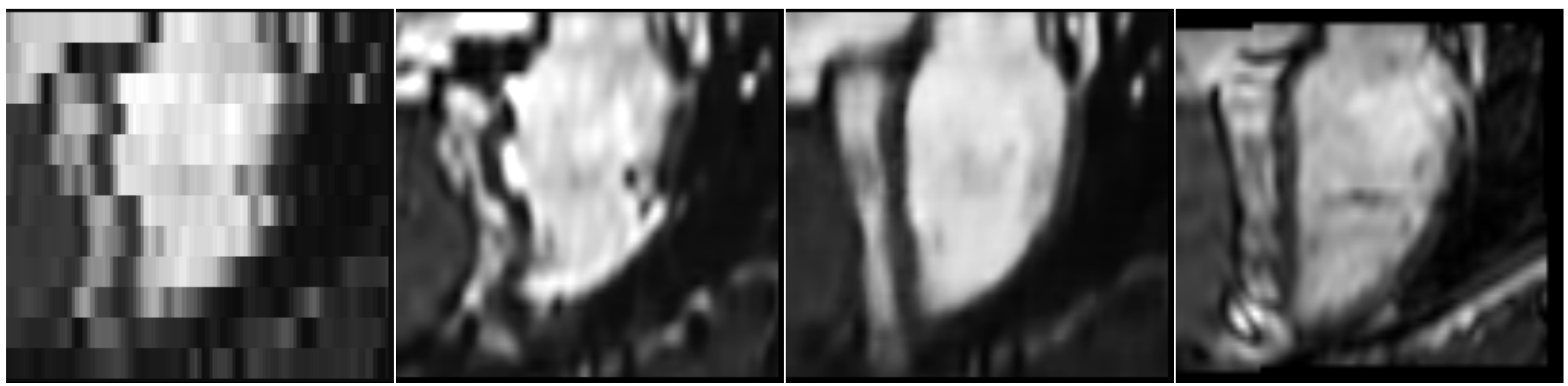}
	\caption{Image super-resolution (SR) results. From left to right, input low resolution MR image, baseline SR approach \cite{oktay2016multi} (no global loss), the proposed anatomically constrained SR model, and the ground-truth high resolution acquisition.}
	\label{fig:resultsMRSR}
\end{figure}

In Table \ref{tab:resultsMRSR}, SSIM and MOS scores for the standard interpolation techniques, SR-CNN, and the proposed ACNN-SR models are provided. In addition to the increased image quality, the ACNN-SR model is computationally more efficient in terms of run-time in comparison to the SR-CNN model \cite{oktay2016multi} by a factor of $5$. This is due to the fact that ACNN-SR performs feature extraction in the low dimensional image space. Furthermore, we investigated the contribution of shape regularisation term in the application of SR, which is visualised in Fig. \ref{fig:resultsMRSR}.

Moreover, we investigated the use of SR as a pre-processing technique for subsequent analysis such as image segmentation, similar to the experiments reported in \cite{oktay2016multi}. In that regard, the proposed SR model and U-Net segmentation models are concatenated to obtain HR segmentation results. However, we observed that the proposed baseline sub-pixel segmentation model (3D-Seg), which merges both SR and segmentation tasks, performs better than the concatenated models. The 3D-Seg approach uses the convolution kernels more efficiently without requiring the model to output a high-dimensional intensity image. For this reason, SR models should be trained by taking into account the final goal and in some cases it's not required to reconstruct a HR intensity image for HR analysis.

\subsection{Learnt Latent Representations and Pathology Classification}
\begin{figure}[t]
	\centering
	\includegraphics[height=67mm,width=.40\textwidth]{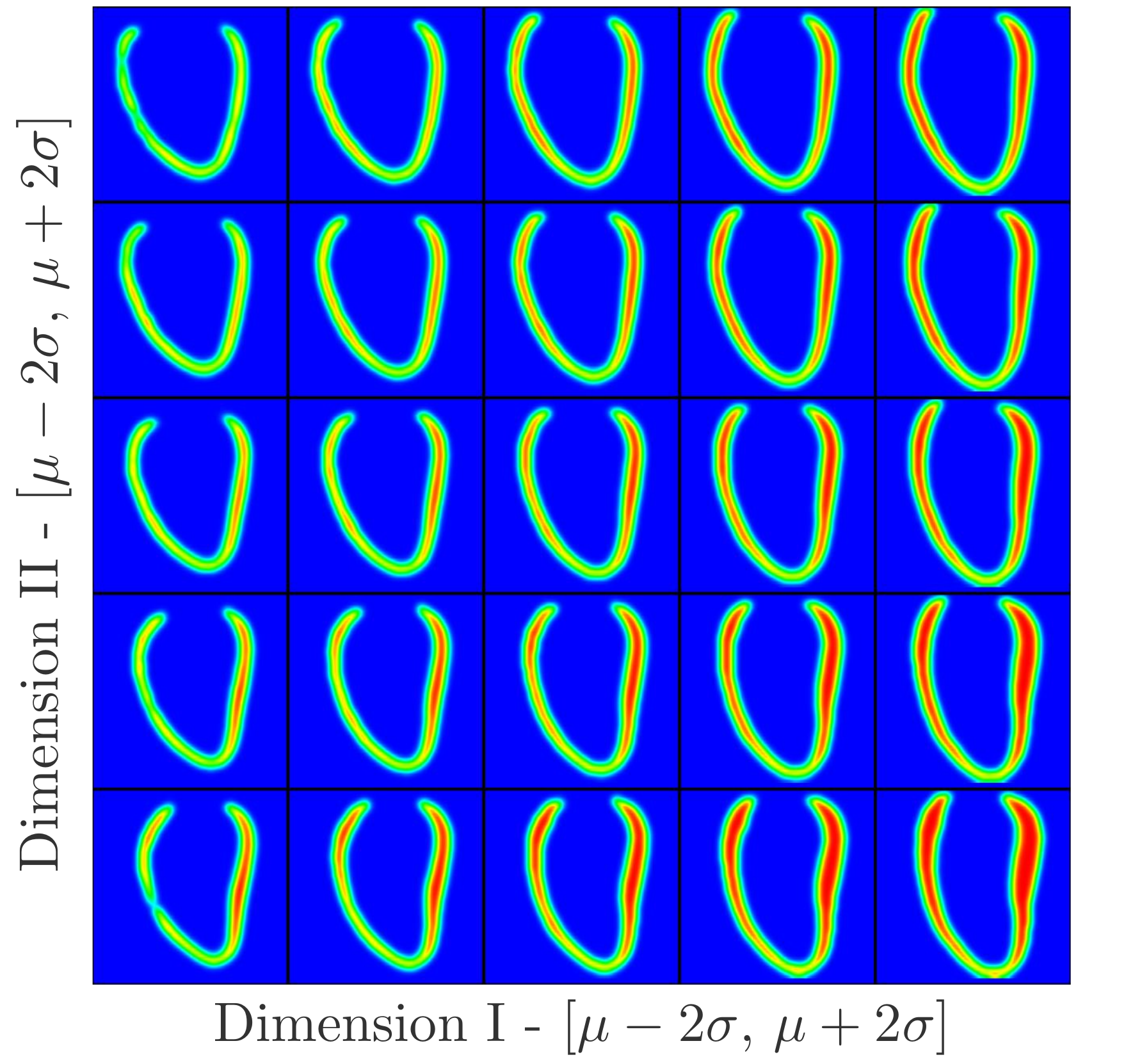}
	\caption{Anatomical variations captured by the latent representations in T-L network (swipe from $\mu-2\sigma$ to $\mu+2\sigma$). Based on our observation, the first and second dimensions capture the variation in the wall thickness of the myocardium (x-axis) and lateral wall of the ventricle (y-axis).}
	\label{fig:meanstdswipe}
\end{figure}

The jointly trained T-L model and its latent representations are analysed and evaluated in the experiment of image pathology classification. This experiment focuses on understanding the information stored in the latent space and also investigates whether they can be used to distinguish healthy subjects from dilated and hypertrophic cardiomyopathy patients. For this, we collected $64$ dimensional codes from segmentation images of the cardiac MR dataset explained in Sec. \ref{sec:ACDC}. Similarly, principal component analysis (PCA) was applied to the same segmentation images (containing LV blood-pool and myocardium labels) to generate $64$ dimensional linear projection of the labels, which requires additional spatial-normalisation prior to linear mapping. The generated codes were then used as features to train an ensemble of decision trees to categorise each image. We used $10$-fold cross-validation on $60$ CMR sequences and obtained $76.6\%$ vs $83.3\%$  accuracy using PCA and T-L codes extracted from ED phase. By including the codes from ES phase, the classification accuracies were improved to $86.6\%$ vs $91.6\%$. This result shows that although the AE and T-L models are not trained with the classification objective, they can still capture anatomical shape variations that are linked to cardiac related pathologies. In particular, we observed that some latent dimensions are more commonly used than others in tree node splits. By sampling codes from the latent space across these dimensions, we observed that the network captures the variation in wall thickness and blood pool size as shown in Fig. \ref{fig:meanstdswipe}. Since we obtain a regular and smooth latent representation, it is possible to transverse along the latent space and generate LV shapes by interpolating between data points. It is important to note that classification accuracies can be further improved by training the AE and T-L models with a classification objective. Our main goal in this experiment was to understand whether the enforced prior distributions contain anatomical information or they are abstract representations only meaningful to the decoder of the AE. 

\section{Discussion and Conclusion}
In this work, we presented a new image analysis framework that utilises autoencoder (AE) and T-L networks as regularisers to train neural network (NN) models. With this new training objective, at testing time NNs make predictions that are in agreement with the learnt shape models of the underlying anatomy, which are referred as image priors. The experimental results show that the state-of-the-art NN models can benefit from the learnt priors in cases where the images are corrupted and contain artefacts. The proposed regulariser model can be seen as an application-specific training objective. In that regard, our model differentiates from the VGG-Net \cite{simonyan2014very} feature based training objectives \cite{ledig2016photo,johnson2016perceptual}. {\blue VGG features tend to be more general purpose representations that are learnt from ImageNet dataset containing natural images of a large variety of objects. In contrast to this, our AE model is trained solely on cardiac segmentation masks and features are customised to identify anatomical variations observed in the heart chambers. For this reason, we would expect the AE features of the segmentations to be more distinctive and informative.} 

{\blue As an alternative to the proposed framework, label space dependencies could be exploited also through adversarial loss (AL) objective functions. Such approaches have been used successfully in natural image super-resolution (SR) \cite{ledig2016photo} and segmentation \cite{luc2016semantic} tasks. In SR application, AL enables the SR network to hallucinate fine texture detail, and the synthesized HR images appear qualitatively more realistic. However, at the same time the PSNR and SSIM scores are usually worse. For this reason, the authors of \cite{ledig2016photo} have pointed out that adversarial training may not be suitable for medical applications, where the accuracy and fidelity of the visual content more important than the qualitative appearance of the HR images. Moreover, we believe that adversarial training comes} at the expense of less interpretability of the regularisation term and unstable model training behaviour, which still remains an open research problem.

Additionally, in the experiments we demonstrated that the learnt codes can be used as biomarkers for classification of cardiac related pathologies and we analysed the distribution of the learnt latent space. This latent space can be further constrained to be Gaussian distributed by replacing the proposed regularisation model with a variational autoencoder. However, this design choice was not considered in our ACNN framework due to two main reasons: (I) the additional K-L divergence term (constraint) would reduce the representation power of the AE; {\blue thus, the local anatomical variations would not be captured in detail.} (II) A generative AE model is not essential for the regularisation of the proposed segmentation and SR models. {\blue A variational architecture would be useful if it was required to sample random instances from the latent space and reconstruct anatomically meaningful segmentation masks; however, in our framework we are only interested in the anatomy specific AE features for model regularisation.} 

The presented ACNN framework is not only limited to the medical image segmentation and SR tasks but can be extended to other image analysis tasks where prior knowledge can provide model guidance and robustness. In that regard, future research will focus on the application of ACNN to the problems such as human pose estimation, anatomical and facial landmark localisation on partially occluded image data. 

{\blue
\section{Acknowledgements}
This study was supported by an EPSRC programme grant (EP/P001009/1) and by British Heart Foundation, UK, project grant (PG/12/27/29489). The research was also co-funded by the National Institute for Health Research (NIHR) Biomedical Research Centre based at Imperial College Healthcare NHS Trust. The views expressed are those of the author(s) and not necessarily those of the NHS, the EPSRC, the NIHR or the Department of Health.
}

% or


\ifCLASSOPTIONcaptionsoff
  \newpage
\fi



%\bibliographystyle{IEEEtran}

{\small
\bibliographystyle{ieee}
\bibliography{references}
}

\clearpage

\section{Supplementary Material}

\subsubsection{Implementation Details} In this section, we give the details about the meta-parameter choices and sampling strategies. In both the super-resolution (SR) and segmentation (Seg) models, the weight decay regularisation term is weighted by $\lambda_2 = 5 \times 10^{-6}$, and gradient descent learning-rate is fixed to $lr=0.001$. The weight of global priors is chosen experimentally to be $\lambda_1 = 0.01$. Mini batch-size, which is the number of samples used for each back-propagated gradient update, is set to be $8$ samples. The models are trained with full images without a need for patch extraction since the cardiac 2D MR image stack size is relatively smaller compared to the available GPU memory (Nvidia GTX-1080). 

In the SR problem, the through-plane upsampling factor is fixed to $K=5$ and the synthetic low-resolution training samples are generated simply by filtering high-resolution images with a Gaussian blurring kernel ($\sigma = 4.0$ mm) along the through plane direction. The blurring operation is followed by a decimation operator along the same image dimension. In the segmentation problem, the Sorensen-Dice loss was tested in the experiments as an alternative to the cross-entropy loss, yet we observed a degraded performance since the latter is a smoother function. 

\subsubsection{Network Structure} In this section, we give the network structures of the autoencoder (AE) and the predictor, which together build the T-L network. The details are provided in Table \ref{tab:PRarchitecture} and \ref{tab:AEarchitecture}. As can be seen in the tables, residual connections are not used in our models ($10$-$18$ layers) since they do not provide significant accuracy gains for smaller networks as reported in \cite{ledig2016photo}. Additionally, non-linear layers are not applied on the lower dimensional latent representations since that would further constrain the autoencoder. The number of the hidden units ($64$) is chosen experimentally, and optimal configurations can be explored to improve the reconstruction performance. The input layer of the AE model is extended to multi-label segmentation maps through one-hot image representation, where each label is converted into a separate channel at the input. Lastly, in both models (AE and PR) each convolution layer, except the last one, is followed by batch-normalisation for better convergence behaviour. 

\begin{table}[h]
	\parbox{0.49\textwidth}{
		\captionof{table}{Structure of the predictor model: The model maps the input HR intensity image ($120$x$120$x$60$) to the latent space and generates a $64$-dim representation. The size, number, and stride of the learnt convolution (Conv) kernels are provided. The filters operate on different image scales ($S1$-$S4$) and each convolution operation is followed by a non-linear unit (ReLU).}
		\begin{tabular}{@{\extracolsep{1pt}}lccccc@{}}
			& & Size    & Stride    & \# Kernels & Non-linearity\\ \midrule
			\parbox[t]{2mm}{\multirow{2}{*}{\rotatebox[origin=c]{90}{$S1$}}}
			& Conv & (f:3,3,3) & (s:1,1,1) &  (N:32) & ReLU \\
			& Conv & (f:3,3,3) & (s:1,1,1) &  (N:32) & ReLU \\ 
			\midrule 
			\parbox[t]{2mm}{\multirow{2}{*}{\rotatebox[origin=c]{90}{$S2$}}} 
			& Conv & (f:3,3,3) & (s:2,2,2) &  (N:64) & ReLU \\
			& Conv & (f:3,3,3) & (s:1,1,1) &  (N:64) & ReLU \\
			\midrule 
			\parbox[t]{2mm}{\multirow{2}{*}{\rotatebox[origin=c]{90}{$S3$}}}
			& Conv & (f:3,3,3) & (s:2,2,2) &  (N:128)& ReLU \\
			& Conv & (f:3,3,3) & (s:1,1,1) &  (N:128)& ReLU \\
			\midrule 
			\parbox[t]{2mm}{\multirow{3}{*}{\rotatebox[origin=c]{90}{$S4$}}}
			& Conv & (f:3,3,3) & (s:2,2,2) &  (N:256)& ReLU \\
			& Conv & (f:3,3,3) & (s:1,1,1) &  (N:1)  & ReLU \\
			& FC 	 & -         &  -        &  (N:64) & None \\
			\bottomrule
		\end{tabular}
		\label{tab:PRarchitecture}
	}
\end{table}

\begin{table}[h!]
 	\parbox{0.49\textwidth}{
 		\captionof{table}{Structure of the autoencoder (AE) model: The encoder part maps the given input segmentation map ($120$x$120$x$60$) to the latent space through convolution (Conv) and fully-connected (FC) layers. The decoder part recovers the input from the low-dimensional representation and outputs a segmentation map ($120$x$120$x$60$). The size, number, and stride of the learnt convolution (Conv) kernels are provided. The filters operate on different image scales ($S1$-$S4$) and each convolution operation is followed by a non-linear unit (ReLU).}

 		\begin{tabular}{@{\extracolsep{1pt}}lccccc@{}}
 			     & & Kernel    & Stride    & \# Kernels & NonLin\\ \midrule
 			\parbox[t]{2mm}{\multirow{2}{*}{\rotatebox[origin=c]{90}{$S1$}}}
 			& Conv & (f:3,3,3) & (s:2,2,1) &  (N:16) & ReLU \\
			& Conv & (f:3,3,3) & (s:1,1,1) &  (N:16) & ReLU \\ 
			\midrule 
			\parbox[t]{2mm}{\multirow{2}{*}{\rotatebox[origin=c]{90}{$S2$}}} 
			& Conv & (f:3,3,3) & (s:2,2,2) &  (N:32) & ReLU \\
			& Conv & (f:3,3,3) & (s:1,1,1) &  (N:32) & ReLU \\
			\midrule 
			\parbox[t]{2mm}{\multirow{2}{*}{\rotatebox[origin=c]{90}{$S3$}}}
			& Conv & (f:3,3,3) & (s:2,2,2) &  (N:64) & ReLU \\
			& Conv & (f:3,3,3) & (s:1,1,1) &  (N:64) & ReLU \\
			\midrule 
			\parbox[t]{2mm}{\multirow{1}{*}{\rotatebox[origin=c]{90}{$S4$}}}
			& Conv & (f:3,3,3) & (s:3,3,3) &  (N:1)  & ReLU \\
			\midrule 
			\parbox[t]{2mm}{\multirow{2}{*}{\rotatebox[origin=c]{90}{$HC$}}}
			& FC   & -         &  -        &  (N:64) & None \\
			& FC   & -         &  -        &  (N:125) & ReLU \\
			\midrule
			\parbox[t]{2mm}{\multirow{2}{*}{\rotatebox[origin=c]{90}{$S4$}}}
			& Deconv & (f:7,7,7) & (s:3,3,3) &  (N:64) & ReLU \\
			& Conv   & (f:3,3,3) & (s:1,1,1) &  (N:64) & ReLU \\
			\midrule 
			\parbox[t]{2mm}{\multirow{2}{*}{\rotatebox[origin=c]{90}{$S3$}}}
			& Deconv & (f:4,4,4) & (s:2,2,2) &  (N:32) & ReLU \\
			& Conv   & (f:3,3,3) & (s:1,1,1) &  (N:32) & ReLU \\
			\midrule 
			\parbox[t]{2mm}{\multirow{2}{*}{\rotatebox[origin=c]{90}{$S2$}}}
			& Deconv & (f:4,4,4) & (s:2,2,2) &  (N:16) & ReLU \\
			& Conv   & (f:3,3,3) & (s:1,1,1) &  (N:16) & ReLU \\
			\midrule 
			\parbox[t]{2mm}{\multirow{2}{*}{\rotatebox[origin=c]{90}{$S1$}}}
			& Deconv & (f:4,4,1) & (s:2,2,1) &  (N:16) & ReLU \\
			& Conv   & (f:3,3,3) & (s:1,1,1) &  (N:3)  & None \\
			
 			\bottomrule
 		\end{tabular}
 		\label{tab:AEarchitecture}
 	}
\end{table}

\end{document}